\documentclass{egpubl}
\usepackage{pg2021}
\usepackage{xspace}
\usepackage{amsmath}
\usepackage{amssymb}
\usepackage{xcolor}
\usepackage{soul}

\SpecialIssuePaper         

\usepackage[T1]{fontenc}
\usepackage{dfadobe}  

\usepackage{cite}  
\BibtexOrBiblatex
\electronicVersion
\PrintedOrElectronic
\ifpdf \usepackage[pdftex]{graphicx} \pdfcompresslevel=9
\else \usepackage[dvips]{graphicx} \fi

\usepackage{egweblnk} 

\usepackage{comment}

\usepackage[ruled,vlined]{algorithm2e}
\usepackage{makecell}

\newcommand{\maps}{guidance images\xspace}

\newcommand{\maptosat}{\emph{map2sat}\xspace}
\newcommand{\seams}{\emph{seam2cont}\xspace}

\newcommand{\name}{SSS\xspace}
\newcommand{\nameLong}{Seamless Satellite-image Synthesis\xspace}

 \newcommand{\bb}{\iffalse}

\newcommand{\red}[1]{\textcolor[rgb]{0.,0.,0.}{#1}}
\newcommand{\reds}[1]{\textcolor[rgb]{0.,0.,0.}{}}
\newcommand{\redr}[2]{\textcolor[rgb]{0.,0.,0.}{#2}}

\title [\nameLong]{\nameLong \xspace}

\author[J. Zhu \& T. Kelly]
{\parbox{\textwidth}{\centering Jialin Zhu\orcid{0000-0002-1826-6566}
        and Tom Kelly\orcid{0000-0002-6575-3682} 
        }
        \\
{\parbox{\textwidth}{\centering University of Leeds}
}
}

\usepackage[switch]{lineno}

\begin{document}

\teaser{
     \def\svgwidth{\linewidth}
\begingroup%
  \makeatletter%
  \providecommand\color[2][]{%
    \errmessage{(Inkscape) Color is used for the text in Inkscape, but the package 'color.sty' is not loaded}%
    \renewcommand\color[2][]{}%
  }%
  \providecommand\transparent[1]{%
    \errmessage{(Inkscape) Transparency is used (non-zero) for the text in Inkscape, but the package 'transparent.sty' is not loaded}%
    \renewcommand\transparent[1]{}%
  }%
  \providecommand\rotatebox[2]{#2}%
  \newcommand*\fsize{\dimexpr\f@size pt\relax}%
  \newcommand*\lineheight[1]{\fontsize{\fsize}{#1\fsize}\selectfont}%
  \ifx\svgwidth\undefined%
    \setlength{\unitlength}{362.60371651bp}%
    \ifx\svgscale\undefined%
      \relax%
    \else%
      \setlength{\unitlength}{\unitlength * \real{\svgscale}}%
    \fi%
  \else%
    \setlength{\unitlength}{\svgwidth}%
  \fi%
  \global\let\svgwidth\undefined%
  \global\let\svgscale\undefined%
  \makeatother%
  \begin{picture}(1,0.23440249)%
    \lineheight{1}%
    \setlength\tabcolsep{0pt}%
    \put(0,0){\includegraphics[width=\unitlength,page=1]{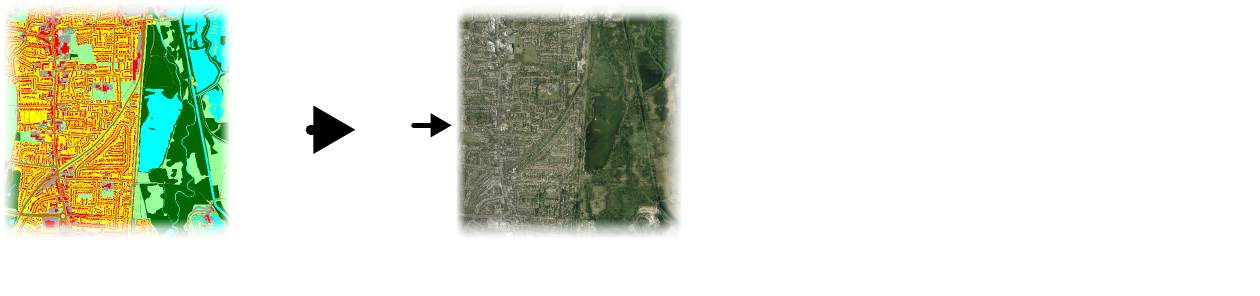}}%
    \put(0.09199004,0.00683364){\color[rgb]{0,0,0}\makebox(0,0)[t]{\lineheight{1.25}\smash{\begin{tabular}[t]{c}input\end{tabular}}}}%
    \put(0.45193568,0.00610308){\color[rgb]{0,0,0}\makebox(0,0)[t]{\lineheight{1.25}\smash{\begin{tabular}[t]{c}$y'^{1}$\end{tabular}}}}%
    \put(0,0){\includegraphics[width=\unitlength,page=2]{teaser.pdf}}%
    \put(0.27576403,0.1287393){\color[rgb]{0,0,0}\makebox(0,0)[t]{\lineheight{1.25}\smash{\begin{tabular}[t]{c}\name{}\end{tabular}}}}%
    \put(0,0){\includegraphics[width=\unitlength,page=3]{teaser.pdf}}%
    \put(0.61294974,0.00581825){\color[rgb]{0,0,0}\makebox(0,0)[t]{\lineheight{1.25}\smash{\begin{tabular}[t]{c}$y'^{2}$\end{tabular}}}}%
    \put(0.72724154,0.00659533){\color[rgb]{0,0,0}\makebox(0,0)[t]{\lineheight{1.25}\smash{\begin{tabular}[t]{c}$y'^{3}$\end{tabular}}}}%
    \put(0.83842731,0.00717746){\color[rgb]{0,0,0}\makebox(0,0)[t]{\lineheight{1.25}\smash{\begin{tabular}[t]{c}$y'^{3}$\end{tabular}}}}%
    \put(0.94903112,0.00659533){\color[rgb]{0,0,0}\makebox(0,0)[t]{\lineheight{1.25}\smash{\begin{tabular}[t]{c}$y'^{3}$\end{tabular}}}}%
    \put(0,0){\includegraphics[width=\unitlength,page=4]{teaser.pdf}}%
  \end{picture}%
\endgroup%

 \centering
  \caption{Our system, \name, takes input vector cartographic data (left), and generates realistic and seamless high-resolution satellite images (right; the image \reds{of tiles }$y'^3$ is 8,192 pixels wide, approximately $67$ megapixels) which are continuous over space and scale. Given the input, \name generates massive spatially seamless images over a range of scales ($y'^1$, $y'^2$, and $y'^3$) by blending the results of \reds{many }lower resolution networks.}
\label{fig:teaser}
}

\maketitle
\begin{abstract}
 We introduce \nameLong (\name), a novel neural architecture to create scale\red{-}and\red{-}space continuous satellite textures from cartographic data. While 2D map data is cheap and easily synthesized, accurate satellite imagery is expensive and often unavailable or out of date. Our approach generates seamless textures over arbitrarily large spatial extents which are consistent through scale-space. To overcome tile size limitations in image-to-image translation approaches, \name learns to remove seams between tiled images in a semantically meaningful manner. Scale-space continuity is achieved by a hierarchy of networks conditioned on style and cartographic data.
 Our qualitative and quantitative evaluations show that our system improves over the state-of-the-art in several key areas. We show applications to texturing procedurally generation maps and interactive satellite image manipulation.


\end{abstract}  

\linespread{0.95}
\section{Introduction}
\label{sec:intro}

Satellite images have revolutionized the way we visualise our environment. Massive, high-resolution, satellite images are essential to many modern endeavors including realistic virtual environments and regional- or city-planning. However, these images can be prohibitively expensive, out of date, or of low quality. Further, continuous satellite images covering large areas are often unavailable due to limited sensor resolution or orbital mechanics. In contrast, cartographic (map) information is frequently accessible and available over massive continuous scales; further, it can be easily manually edited or synthesized. In this paper we address the problem of generating satellite images from map data at arbitrarily large resolutions.

Convolutional neural networks are state of the art for image to image translation; such networks can translate from cartographic images to satellite images. However, these networks are unable to generate outputs of arbitrary areas -- the resolution of these synthetic satellite image tiles remains limited by available memory and compute. Arranging many such small tiles to cover a large map results in inconsistencies (seams) between the tiles. Further, if a larger number of tiles are joined together and viewed at multiple scales, the result lacks continuity and style coordination.

Inspired by convolutional image-to-image translation systems such as Pix2Pix~\cite{pix2pix} and SPADE~\cite{park2019semantic}, \emph{\nameLong} (\name,  Figure~\ref{fig:teaser}) \reds{synthesizes}\red{creates} satellite image \red{tiles} by conditioning generation on \reds{tiles of }cartographic \redr{images}{data}. These tiles are of fixed resolution (256 $\times$ 256 pixels in our implementation) and exhibit good continuity internally, but would contain spatial and scale artifacts if na\"ively tiled \redr{to cover}{over} a larger area. \emph{Spatial discontinuities} between adjacent tiles are caused by \redr{a lack of tile content continuity for properties not determined by the cartographic data}{ a lack of coordination between network evaluations,} resulting in lack of continuity for properties not determined by the cartographic input (e.g., color of building roof or species of tree). We introduce a network which removes such spatial discontinuities by leveraging learned semantic domain knowledge as well as the cartographic data. \emph{Scale discontinuities} between a larger set of synthetic satellite tiles arise from unnatural style variations over \redr{a }{larger areas (e.g., distribution of cars throughout a city or different camera sensors)}. To address this, \name uses a hierarchy of networks synchronized by color guidance.

The \name system contains three components -- two convolutional neural networks to generate tiles of different resolutions (\maptosat), and remove seams (\seams) between those tiles, as well as an interface allowing the generated textures to be explored interactively. \name contributes:
\begin{itemize}
    \item A neural approach to generate massive scale-and-space continuous images conditioned on cartographic data.
    \item A novel architecture to train and evaluate image-to-image generation networks at high resolutions with reduced memory use. 
    \item An application to texture procedurally generated cartographic maps -- allowing the the synthesis of textures for entirely novel areas \red{following user style guidance}.
    \item An application to interactive exploration of endless satellite images generated on-the-fly from cartographic data.
\end{itemize}

\red{The textures generated by \name have many applications, including interactive city planning, virtual environments, and procedural modeling. During city planning, the textures may help stakeholders (e.g., homeowners, planners, and motorists) understand proposed developments. For example, sit-down sessions in which stakeholders interactively edit a map and are shown birds-eye-view (satellite) ``photos'' within seconds. Further, virtual environments may be constructed using \name's output (e.g., a flight simulator) to texture massive terrains on-the-fly, or the streets and open areas of urban procedural models~\cite{Parish:2001:PMC}.}

In the remainder of this work we will discuss the corpus surrounding our work in Section \ref{sec:related}, introduce our system in Section~\ref{sec:overview}, give the details of the neural architectures and their coordination in section~\ref{sec:method} before evaluating our results qualitatively and quantitatively in section~\ref{sec:results}. \reds{\emph{Upon publication our source code, retrained network weights, and dataset pre-processing tools will be released.}} \red{Source code and weights for our system are available online at} \url{https://github.com/Misaliet/Seamless-Satellite-image-Synthesis}.

\section{Related Work}
\label{sec:related}

\begin{figure*}[t!]
    \centering
    \def\svgwidth{\linewidth}
    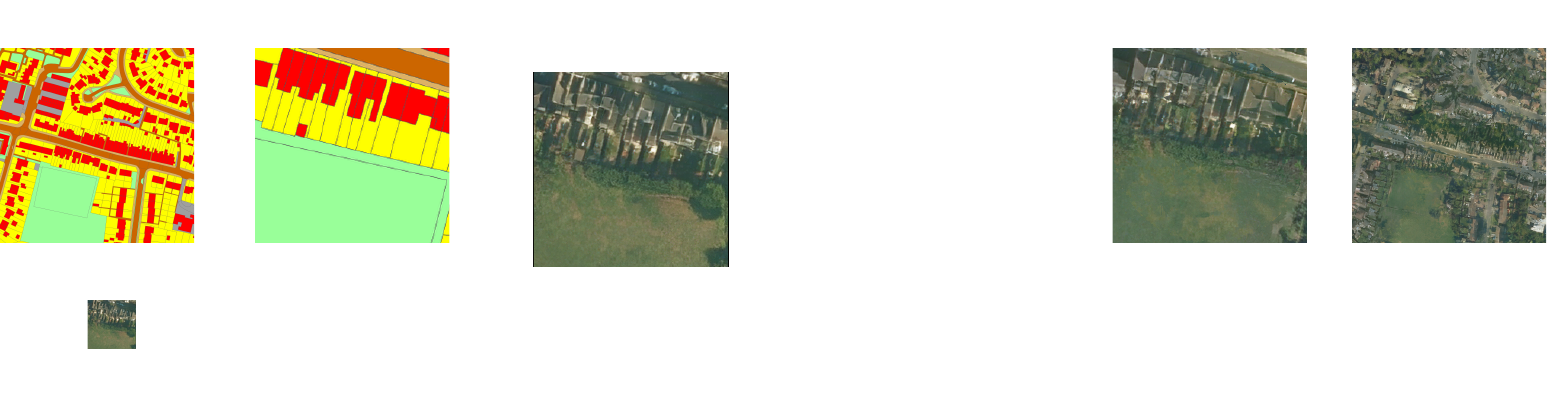
    \caption{\name uses a hierarchy of networks for different scales. A pair of networks is shown above, which process input cartographic maps to large seamless images at a single scale level, $z$. First, a vector map of arbitrary extent is processed to create an image tile of map data. The \maptosat network then performs a translation to the satellite texture domain, before the \seams network removes visible tilling artifacts to create a seamless image and the result is combined with neighboring tiles to create an image of arbitrary resolution (right). Subsequent, smaller scale, \maptosat networks use previous outputs as scale-guidance (bottom left).}
    \label{fig:overview}
\end{figure*}

\textbf{Texture synthesis} is the generation of texture images from exemplars or lower dimension representations. We refer the reader to Wei et al.~\cite{SynthesisPauly} for an overview. Non-neural approaches from the AI Winter create images with local and stationary properties using techniques such as Markov Random Fields to generate pixels~\cite{efros1999texture,WeiTree} or patches~\cite{LiangPatch, efros2001image} from an exemplar image. Variations on these themes generate larger, more varied textures~\cite{darabi2012image}. Procedural languages for texturing~\cite{muller2007image} are well adapted for urban domains and can be used to guide synthesis~\cite{guehl2020semi}.

With the advent of \textbf{Convolutional Neural Networks (CNNs)}, texture synthesis has increasingly become data-driven\reds{technique}, creating \redr{textures}{images} by gradient descent in feature space~\cite{gatys2015texture} or with Generative Adversarial Networks (GANs)~\cite{ZhouTextureSynthesis}. 
Another approach to texture generation is to transform an input image to a different style or domain while retaining the structure. Such \emph{image-to-image} translation is implemented \redr{in}{by} the influential UNet architecture~\cite{ronneberger2015u} which passes the input through an ``hourglass'' shaped network of convolutional layers. This allows the network to aggregate and coordinate features within a\reds{ hourglass} \emph{bottleneck}. UNets have been applied to generate satellite images~\cite{pix2pix} from map data, however the resolution of the output tiles was limited \red{by available memory}.
Image-to-image networks typically disentangle the structure from style to enable style transfer~\cite{gatys2016image}. These have been applied to generate terrains~\cite{guerin2017interactive} and \red{within} chained sequences~\cite{kelly2018frankengan} for architectural texture pipelines.

\textbf{Label-to-image translation} is a variation of image-to-image translation; the input is a label-map with each pixel having a specific class, such as cartographic data containing, streets, buildings, and vegetation. \redr{One approach is}{Again, we note the application of} hourglass networks~\cite{pix2pixhd}. We build on Semantic Image Synthesis with Spatially-Adaptive Normalization
(SPADE)~\cite{park2019semantic} which uses a simpler V-shaped architecture with a bottleneck at the start, rather than center, of the network. From a latent code, successive \emph {SPADE Residual Blocks} generate feature-maps of increasing resolution. Within these blocks, a SPADE unit introduces learned per-label style information using Spatially-Adaptive Normalization.

\textbf{Texture blending or inpainting} approaches give a way to combine multiple smaller images into a larger one. These may match small image \red{patches~\cite{LiangPatch, efros2001image,Barnes:2009:PAR}, or minimize a function over a grid of pixels using seam energy minimization~\cite{seamCarvingResize,kwatra2003graphcut, rother2004grabcut}, exemplars~\cite{criminisi2003object}}, or a Poisson formulation~\cite{perez2003poisson}. 
Although such local approaches produce excellent pixel-level results, they do not contain the higher, semantic-level knowledge to blend textures and images with complex structures. We may overcome this limitation with user input~\cite{agarwala2004interactive}, large datasets~\cite{HaysSceneCompletion}, and distributions learned by neural architectures~\cite{zhan2019spatial,zhang2020deep, Bau_semantic}. These learn distributions for the blended area using techniques such as guided Poisson constraints~\cite{wu2019gp} and symmetry~\cite{deng2018uv}.
Domain-specific image blending has been studied for domains such as microscopy~\cite{legesse2015seamless}, fa\c{c}ade texturing~\cite{kelly2018frankengan}, and satellite generation. TileGAN~\cite{fruhstuck2019tilegan} searches a library of latent encodings of adjacent micro-tiles to create seamless images, rather than learning a domain specific blending function.

In contrast to relatively recent satellite images, applications of \textbf{cartography} date back to antiquity. Today, there are high-quality, publicly available maps of much of the world~\cite{osm, digimap}. We may also generate maps with a variety of techniques, including procedural systems; \redr{S}{s}ee Vanegas et al.~\cite{vanegas2010modelling} for an overview. In particular, we may use \reds{a }rule and parameter controlled geometric algorithms to recreate the distribution of streets~\cite{chen2008interactive}, buildings~\cite{Parish:2001:PMC}, \red{and }city parcels~\cite{vanegas2012procedural}. These procedural systems can generate massive quantities of 3D city geometry from random seeds; \red{such geometry} can be used to generate cartographic maps by projecting the 3D meshes to 2D images and applying a constant color for each category (e.g., \redr{all buildings are red}{red buildings}). Procedurally generated cartography has the advantage that it can be easily edited with a user-interface or created from databases.

\section{Overview}
\label{sec:overview}

We propose a novel method for map-to-satellite-image translation over arbitrarily large areas. Two neural networks work together to first generate satellite image tiles from map data, and then remove seams and discontinuities from these images. To generate an area, \reds{the}\name (Figure~\ref{fig:overview}) first generates two overlapping sets of satellite image tiles \reds{for the largest scale level }with the \maptosat network. \redr{The next network}{Then the} \seams \red{network} removes the discontinuities and seams between adjacent tiles by learning to \redr{use}{generate} a mask to blend the overlapping satellite images. Joining together such tiles results in a single seamless image \redr{at}{with} a large resolution and scale.

To achieve our goals, we solve problems such as the training of a mask generator to remove seams while realistically blending tiles, the coordination of networks \reds{which run }over different areas to create continuous results in scale-space, and reducing the memory usage for large image synthesis networks.

\section{Method}
\label{sec:method}

\begin{figure}
    \centering
    \def\svgwidth{\linewidth}
\begingroup%
  \makeatletter%
  \providecommand\color[2][]{%
    \errmessage{(Inkscape) Color is used for the text in Inkscape, but the package 'color.sty' is not loaded}%
    \renewcommand\color[2][]{}%
  }%
  \providecommand\transparent[1]{%
    \errmessage{(Inkscape) Transparency is used (non-zero) for the text in Inkscape, but the package 'transparent.sty' is not loaded}%
    \renewcommand\transparent[1]{}%
  }%
  \providecommand\rotatebox[2]{#2}%
  \newcommand*\fsize{\dimexpr\f@size pt\relax}%
  \newcommand*\lineheight[1]{\fontsize{\fsize}{#1\fsize}\selectfont}%
  \ifx\svgwidth\undefined%
    \setlength{\unitlength}{324.78940228bp}%
    \ifx\svgscale\undefined%
      \relax%
    \else%
      \setlength{\unitlength}{\unitlength * \real{\svgscale}}%
    \fi%
  \else%
    \setlength{\unitlength}{\svgwidth}%
  \fi%
  \global\let\svgwidth\undefined%
  \global\let\svgscale\undefined%
  \makeatother%
  \begin{picture}(1,0.49668898)%
    \lineheight{1}%
    \setlength\tabcolsep{0pt}%
    \put(0,0){\includegraphics[width=\unitlength,page=1]{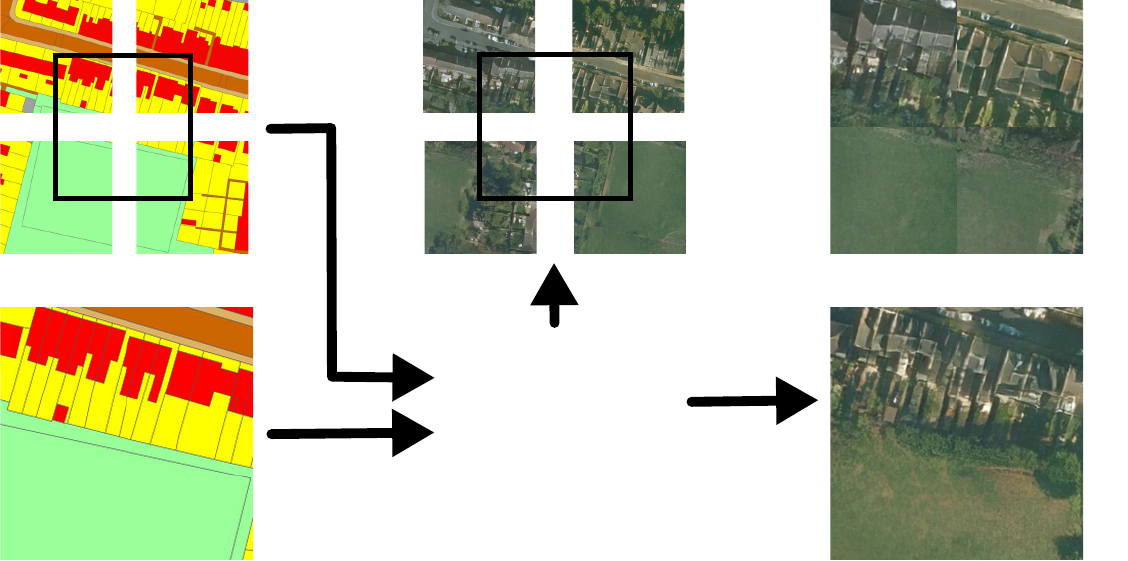}}%
    \put(0.97946899,0.10180414){\color[rgb]{0,0,0}\makebox(0,0)[t]{\lineheight{1.25}\smash{\begin{tabular}[t]{c}$t$\end{tabular}}}}%
    \put(0.9806296,0.37615157){\color[rgb]{0,0,0}\makebox(0,0)[t]{\lineheight{1.25}\smash{\begin{tabular}[t]{c}$s$\end{tabular}}}}%
    \put(0,0){\includegraphics[width=\unitlength,page=2]{st.pdf}}%
    \put(0.49537611,0.13143203){\color[rgb]{0,0,0}\makebox(0,0)[t]{\lineheight{1.25}\smash{\begin{tabular}[t]{c}\maptosat\end{tabular}}}}%
    \put(0,0){\includegraphics[width=\unitlength,page=3]{st.pdf}}%
  \end{picture}%
\endgroup%

    \caption{Creation of \seams input tiles. We obtain one image with seams, $s$ and one without seams, $t$, from the \maptosat network run over map data (left). Note that the corners of $s$ are seamless against adjacent $s$ tiles as they come from the same \maptosat execution.}
    \label{fig:st}
\end{figure}

\name uses convolution neural networks (CNNs) to learn to synthesize satellite image by learning over \reds{representative }datasets. We prepare \red{a dataset of} paired map and satellite images for each tile at a fixed resolution of 256 $\times$ 256 pixels. \reds{The map data is processed from a}Vector cartographic \reds{data description }data \red{is processed to image} tiles -- each \red{GeoPackage~\cite{geopackage}} file contains a list of polygons, each with one of 13 semantic class labels (e.g., \emph{building, road or track}, or \emph{natural environment}). The polygons are ordered canonically and \reds{then }rendered into an image, assigning every a pixel a label. The \red{corresponding} satellite images have 8-bit color depth, a resolution \redr{0.25cm/pixel}{0.25 meters per pixel}, and metadata containing their  \redr{coordinates}{location and extent} in the local Coordinate Reference System; we use this transform to align them to the map data.

The paired dataset is used to train \redr{an}{our} \red{image-to-image} translation network, \maptosat, based on the SPADE\cite{park2019semantic} architecture. The network learns to translate our semantic cartographic label image \red{tiles} into corresponding satellite image\reds{over fixed resolution} tiles.  However, to synthesize massive areas two \reds{significant }problems must be solved.

The first problem is maintaining a realistic style over different scales of tiled areas: \emph{scale-space continuity}. If our tiles are too large, we are able to synthesize a large area, but the limited fixed resolution provides poor detail (\redr{S}{s}ee additional materials \emph{variants\_8k\_images\slash large\_scale\_only.jpg}). On the other hand, if the tiles are too small, we see unrealistic variation over the tiled image -- this may be too little variation, resulting in a homogeneous image (e.g., very similar trees over an entire city); see \reds{See }additional materials \emph{variants\_8k\_images\slash small\_scale\_only.jpg} -- or too much variation, leading to a chaotic swings in is variation between adjacent tiles (e.g., roads suddenly changing color). \red{Additionally, }our interactive system allows users to explore a dynamically generated satellite image at a range of scales; \redr{For}{in} this application, it is necessary to generate imagery at a given scale without overly onerous number of generator network applications. \name uses a scale-space hierarchy of generator networks to create realistic variation over areas at multiple scales (Section~\ref{sec:seams}). By evaluating the hierarchy tiles lazily, the interactive system is able to quickly evaluate tiles at arbitrary scales. 

The second problem is that simple tiling of the synthetic satellite images to cover a large area does not produce natural images, but creates an image with seams; state-of-the-art CNN architectures do not provide \emph{spatial-continuity} between adjacent tiles. This is largely because of their reliance on \red{spatially} narrow bottleneck layers, which allow tile-wide coordination of features \redr{, with the consequence of }{at the cost of} limiting maximum tile size. Adding additional layers to increase the tile size becomes prohibitively expensive \reds{during training }due to the \redr{image size}{memory requirements of these large layers}. \reds{An alternative are ``fully-convolutional'' architectures without per-tile bottlenecks, but without such efficient coordination, these do not achieve the same accuracy or realism.} \red{Another source of seams is the padding~\cite{alsallakh2021mind} around layer boundaries, where synthetic values (i.e., zero or flipped) are used to pad convolution input; these introduce seam artifacts around the perimeter of network outputs. We address this spatial-continuity problem in the following section.}

\subsection{Spatial Continuity: \seams Network}
\label{sec:seams}

The limited resolution of CNN image synthesis techniques, such as those we use in the \maptosat network, leads to seam discontinuities when we join multiple tiles together. Here we introduce the \seams network which removes the seams\redr{ to allow}, allowing tiling \redr{ to cover}{ over} an arbitrarily large area. Given tiles generated by our image-to-image network we introduce a novel architecture to blend pixels from these existing tiles to create a seam-free image. \red{We found blending (masking) a sufficient approach to seam removal as the tile content is well aligned by strong cartographic conditioning, and blending over a wide area allows the network to identify semantic plausible locations for each transition.}

For the area of one tile $i$, we use two independent, overlapping layers of \maptosat evaluation to synthesize tiles $s_i$ and $t_i$. The areas of $s_i$ and $t_i$ are offset so for each tile $i$, there is a seam-free option\reds{in either $s_i$ or $t_i$}, as Figure ~\ref{fig:st}. Our goal is to blend $s_i$ and $t_i$ such that the result has no internal seams (as $t_i$), but also \red{has} no seams with neighboring tiles (as $s_i$).
The quarters of $s_i$ are continuous across the tile boundary because they have come from the same output tile of a single \maptosat evaluation. Thus, the problem of removing seams is transformed to \red{a} masking \red{(image blending)} problem -- where the mask predominantly selects pixels from $t_i$ in the center and from $s_i$ in the periphery\reds{ of the result}. In-between it should select pixels which maximize realism.  We propose a method to learn to generate the mask, $m$, which selects how to blend $s_i$ and $t_i$. By learning a mask generator, \seams can learn the semantic meanings of the inputs, producing higher quality results than optimizing for gradient~\cite{kwatra2003graphcut} or search~\cite{fruhstuck2019tilegan}.

For one tile area $i$, \seams takes \reds{as the conditional } the map image $x_i$ (including label and instance data), the generated satellite image with seams ($s_i$), and the generated satellite image without seams ($t_i$). The output of the \seams generator is a mask, $m_i$, which is used to blend the input satellite images. The output of the system is seamlessly tillable satellite image, $y'_i$, with a feature distribution similar to the seamless ground-truth images.

\begin{figure*}
    \centering
    \def\svgwidth{\linewidth}
    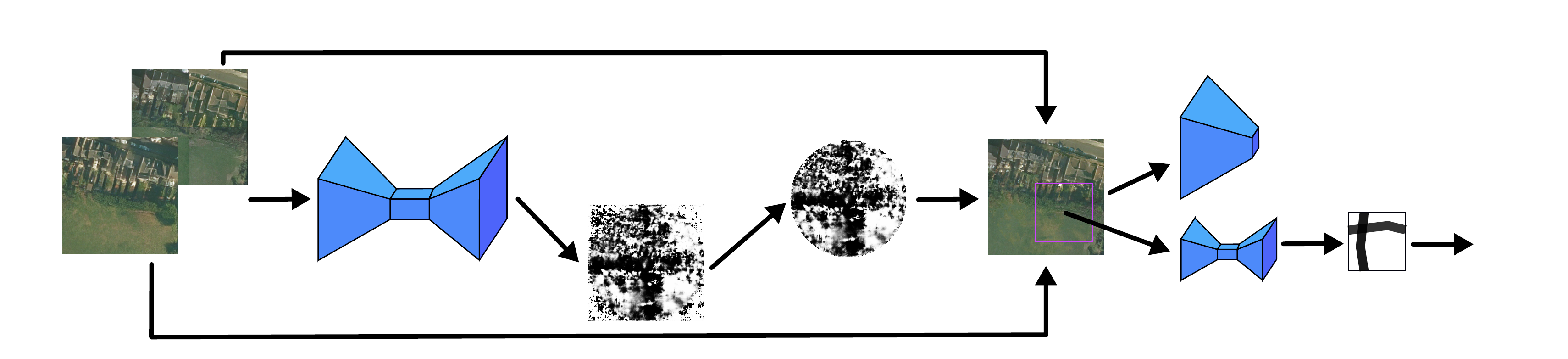
    \caption{The data-flow when testing the \seams network (black arrows). The inputs are map label images $x_i$, as well as generated satellite images $s_i$ and $t_i$. $M$ generates a mask, $m_i$ which is multiplied by a fixed circle image $c$ to create $m'_i$. This final mask blends  $s_i$ and $t_i$ to create our spatially continuous seamless image $y'_i$. Cyan arrows: the two discriminators contribute losses when training $M$.}
    \label{fig:seams}
\end{figure*}

Figure~\ref{fig:seams} shows the architecture of the \seams network. We use a UNet\cite{ronneberger2015u, zhu2017multimodal} as our generator $M$ to create a mask, $m_i$, from inputs $x_i$, $s_i$ and $t_i$:
        \begin{equation}
            m_i = M(x_i, s_i, t_i) 
        \end{equation}
To evaluate the quality of the generated mask, an output tile, $y'_i$ is blended from $s_i$ and $t_i$ by $B$:
        \begin{equation}
            y'_i = B(s_i, t_i, m_i) = s_i * m_i + t_i * (1 - m_i)
            \label{y'_i}
        \end{equation}

$M$ is a 8 layer UNet which we train with three goals for the blended image $y'_i$: (1) tileable -- with $s_i$ near the borders, (2) realistic, and (3) internally seamless. \redr{To train the mask generator w}{W}e construct a differentiable loss function to achieve these three goals which can be backpropogated to improve $M$ \red{during training}.

To address goal (1) the mask should select  $s_i$ at the edge of the tile to ensure continuity with adjacent tiles. We implement this by forcing the outside part of generated mask $m_i$ to be white (to have a value of $1$) in $F$, guided by a constant circular mask $c$ (Figure~\ref{fig:seams}):
        \begin{equation}
            m'_i = F(m_i) = m_i * (1 - c) + c
        \end{equation}
and we can modify equation \ref{y'_i} to give us the masked output:
        \begin{equation}
            y'_i = B(s_i, t_i, m'_i) = s_i * m'_i + t_i * (1 - m'_i)
        \end{equation}

To support goal (2) we use a discriminator $D_{realism}$, a PatchGAN~\cite{pix2pix}, to reward $M$ for the realism of the masked output.  $D_{realism}$ is trained to discriminate between generated images and those from the ground-truth (without seams). When training $M$, it takes results $y'_i$ and the constant circular mask $c$. These conditions allow the discriminator to locate the output seams~\cite{teterwak2019boundless} and increase accuracy. The loss $D_{realism}$ is then:
        \begin{equation}
            \mathcal{L}_{realism} = \mathbb{E}_{x_i, s_i, t_i}[ -D_{realism}(y'_i, c)]
        \end{equation}
Following the standard GAN architecture, this loss using $D_{realism}$ is used during training to update $M$ to increase realism.

Goal (3) is to remove internal seams from $y'_i$. To achieve this, we train an image-to-image CNN to locate seams and use it as a discriminator, $D_{seams}$. It encourages $M$ to generate a mask $m_i$ which removes seams from the blended result. It is an image-to-image translation network which learns to identify seams in patches of $y'_i$, outputting a grayscale image that is white (value of 1) where there is no seam and black (value of 0) where there is a seam. $D_{seams}$ is used as an adversarial loss when training $M$ by constructing a L2 loss between the discriminator's output and the desired (seamless; value of 1) result. Following CUT~\cite{CUT}, we extract random patches, $P$, from the output of mask to contribute to the loss:
        \begin{equation}
            \mathcal{L}_{seams} =  ||D_{seams}(P(y'_i)) -1||
        \end{equation}
Finally, as Pix2Pix\cite{pix2pix}, we add a L1 loss to the ground-truth in order to improve the training stability:
        \begin{equation}
            \mathcal{L}_{recons} = |y'_i - y_i|
        \end{equation}
The total loss function that is propagated through $M$ during training is then:
        \begin{equation}
            \mathcal{L}_{M} = \lambda_{1}\mathcal{L}_{realism} + \lambda_{2}\mathcal{L}_{seams} +  \lambda_{3}\mathcal{L}_{recons}
        \end{equation}
We train the model with Adam~\cite{DBLP:journals/corr/KingmaB14} with a learning rate of $0.0002$. We \red{may write the} complete \seams network, $S$, which generates creating seamless image, $y'_i$, as:
    \begin{equation}
        S(x_i, s_i, t_i) = B(s_i, t_i, F(M(x_i, s_i, t_i))) = y'_i
    \end{equation}

We train the neural networks $M$, $D_{realism}$, and $D_{seams}$ together, performing one SGD update for each network at every iteration. $D_{realism}$ is trained to discriminate between blended result images ($y'_i$) and image without seams ($t_i$):
    \begin{equation}
        \mathcal{L}_{d-realism} = \mathbb{E}_{x_i, s_i, t_i} [D_{realism}(s_i, c) + (1 - D_{realism}(y'_i, c))]
    \end{equation}
$D_{seams}$ is trained to identify seam locations in images. We find that the simplicity of this image based seams/no-seams formulation complements $D_{realism}$, which works on more complex image space features. We use random patches with $D_{seams}$  as they are smaller and \redr{easier}{faster} to train. Random patches also have non-constant output (with seams in different locations) as they are unaffected by boundary layer padding. We train $D_{seams}$ to discriminate between the output $n$ and a seams mask image showing the position of seams, $l$, with the loss:
    \begin{equation}
        \mathcal{L}_{d-seams} = ||D_{seams}(P(y'_i)) - P(l) || 
    \end{equation}
where $P$ is a particular random crop of resolution 64 $\times$ 64 pixels.

The complete loss function of \seams network that is backpropagated through $M$ when training is:
\begin{equation}
\begin{aligned}
    \mathcal{L}_{total} = \lambda_{1}\mathcal{L}_{realism} + \lambda_{2}\mathcal{L}_{seams} +  \lambda_{3}\mathcal{L}_{recons}\\
    +\lambda_{4}\mathcal{L}_{d-realism} + \lambda_{5}\mathcal{L}_{d-seams}
\end{aligned}
\end{equation}

The training data for \red{the} whole \seams network contains seams images and non-seams images, both of them are generated by \maptosat at different scales. We train \seams network with a dataset of around 3,000 images \reds{of each type}on a Nvidia Titan V GPU for 200 epochs \red{using hyperparameters $\lambda_1=1.0$, $\lambda_2=1.0$, $\lambda_3=10.0$, $\lambda_4=1.0$, and  $\lambda_5=1.0$.}

The primary resolution of all networks in \name is 256 $\times$ 256 pixels. \red{Larger resolutions are} possible, but would require significantly increased memory. This limited resolution allows us to study impact of seam-removal with reasonable compute resources.

\subsection{Scale-Space Continuity: \maptosat Network}
\label{sec:zoom}

\begin{figure}
    \centering
    \def\svgwidth{\linewidth}
\begingroup%
  \makeatletter%
  \providecommand\color[2][]{%
    \errmessage{(Inkscape) Color is used for the text in Inkscape, but the package 'color.sty' is not loaded}%
    \renewcommand\color[2][]{}%
  }%
  \providecommand\transparent[1]{%
    \errmessage{(Inkscape) Transparency is used (non-zero) for the text in Inkscape, but the package 'transparent.sty' is not loaded}%
    \renewcommand\transparent[1]{}%
  }%
  \providecommand\rotatebox[2]{#2}%
  \newcommand*\fsize{\dimexpr\f@size pt\relax}%
  \newcommand*\lineheight[1]{\fontsize{\fsize}{#1\fsize}\selectfont}%
  \ifx\svgwidth\undefined%
    \setlength{\unitlength}{771.13028459bp}%
    \ifx\svgscale\undefined%
      \relax%
    \else%
      \setlength{\unitlength}{\unitlength * \real{\svgscale}}%
    \fi%
  \else%
    \setlength{\unitlength}{\svgwidth}%
  \fi%
  \global\let\svgwidth\undefined%
  \global\let\svgscale\undefined%
  \makeatother%
  \begin{picture}(1,0.26002305)%
    \lineheight{1}%
    \setlength\tabcolsep{0pt}%
    \put(0,0){\includegraphics[width=\unitlength,page=1]{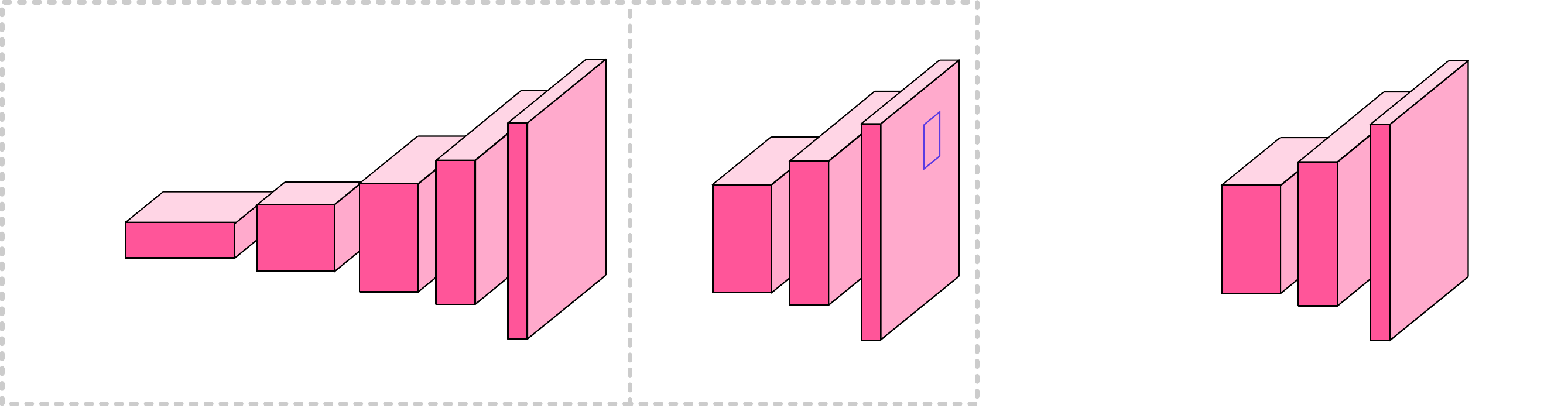}}%
    \put(0.03074972,0.10393978){\color[rgb]{0,0,0}\makebox(0,0)[lt]{\lineheight{1.25}\smash{\begin{tabular}[t]{l}$e$\end{tabular}}}}%
    \put(0,0){\includegraphics[width=\unitlength,page=2]{map2sat.pdf}}%
    \put(0.23347713,0.00957732){\color[rgb]{0,0,0}\makebox(0,0)[t]{\lineheight{1.25}\smash{\begin{tabular}[t]{c}$\maptosat^1$\end{tabular}}}}%
    \put(0.52013794,0.00957237){\color[rgb]{0,0,0}\makebox(0,0)[t]{\lineheight{1.25}\smash{\begin{tabular}[t]{c}$\maptosat^2$\end{tabular}}}}%
    \put(0.8859725,0.00866003){\color[rgb]{0,0,0}\makebox(0,0)[t]{\lineheight{1.25}\smash{\begin{tabular}[t]{c}$\maptosat^{z_{max}}$\end{tabular}}}}%
    \put(0,0){\includegraphics[width=\unitlength,page=3]{map2sat.pdf}}%
    \put(0.67937643,0.14174383){\color[rgb]{0,0,0}\makebox(0,0)[lt]{\lineheight{1.25}\smash{\begin{tabular}[t]{l}...\end{tabular}}}}%
    \put(0,0){\includegraphics[width=\unitlength,page=4]{map2sat.pdf}}%
  \end{picture}%
\endgroup%

    \caption{The \maptosat sub-networks can be trained independently or end-to-end. This figure illustrates the organization of SPADE residual blocks (magenta, Figure~\ref{fig:spade_block}). $\maptosat^1$ is a SPADE network taking latent input $e$. The output of $\maptosat^1$ is cropped (purple) to provide color guidance to the $\maptosat^2$ sub-network. $\maptosat^2$ and later sub-networks have the same architecture.}
    \label{fig:zoom_arch}
\end{figure}

\begin{figure}
    \centering
    \def\svgwidth{\linewidth}
    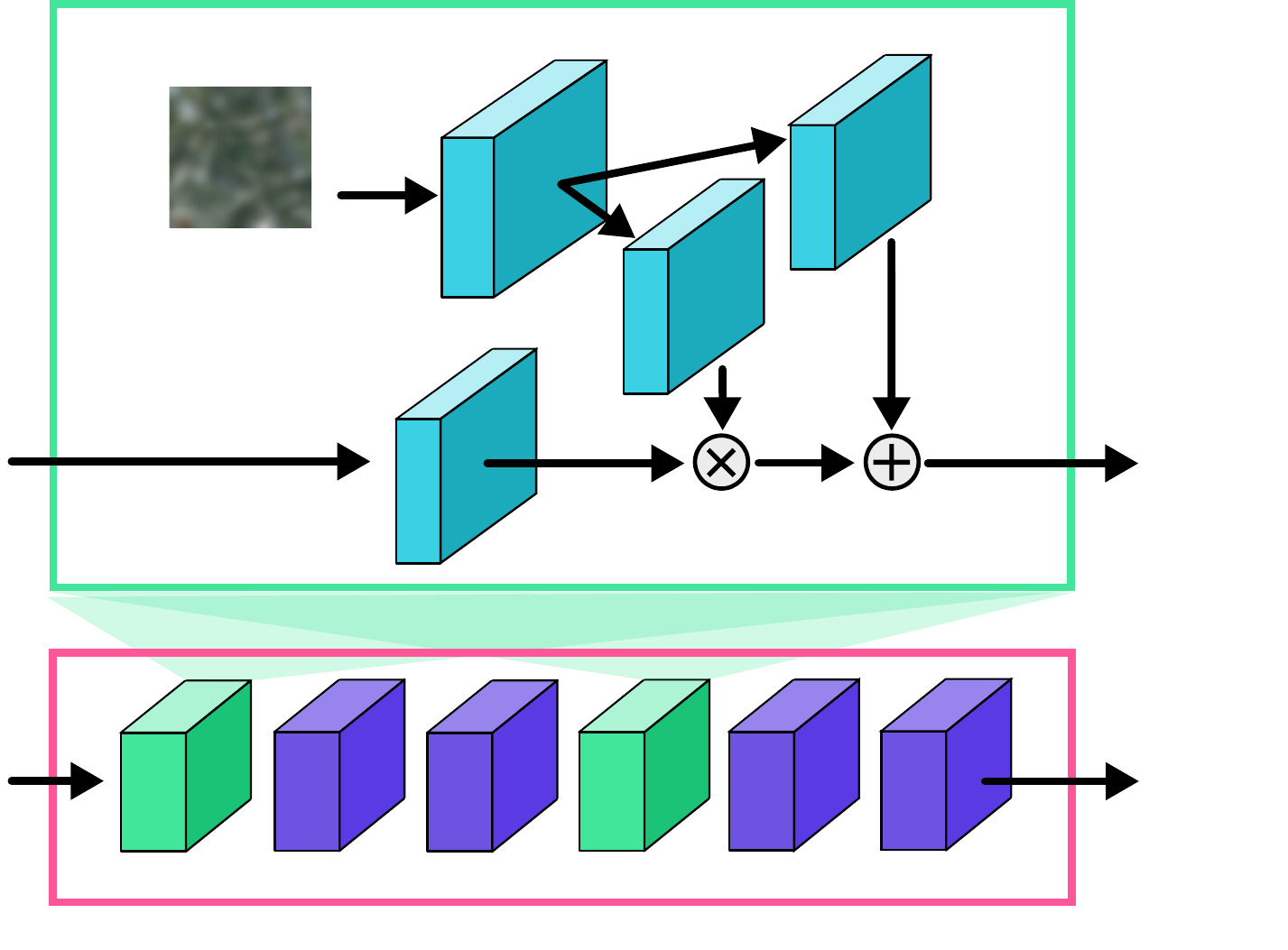
    \caption{Top (green): Extending the original SPADE block~\cite{park2019semantic}, we concatenate the RGB color guidance from the previous $\maptosat$ scale and the map image. The concatenated tensor is used to learn convolutions to manipulate the channels after batch-normalisation. Bottom: SPADE residual blocks are combined with ReLU and convolutional layers to create the magenta blocks shown in Figure~\ref{fig:zoom_arch}.}
    \label{fig:spade_block}
\end{figure}

The \seams network allows us to build large composite images from generated tiles. To generate these tiles, we build on \red{SPADE~\cite{park2019semantic} to create}\reds{ generate} realistic \redr{single}{individual} tiles from map label images. These tiles can be assembled into an arbitrarily large image. However, there is no coordination between tiles \red{which are} far apart. To address this scale-space problem, we introduce \reds{the }the \maptosat network to generate synthetic image tiles with scale-space continuity.

To provide consistency at different scales, we would ideally use a single large tile, however such high resolution is prohibitively computationally expensive. Instead, \maptosat uses multiple sub-networks to generate tiles at different \emph{scales}, \red{retaining a fixed \emph{resolution} of 256 $\times$ 256 pixels. For example,} at the largest scale, $z=1$, map image tiles are 1 kilometer wide \red{(3.9 meters per pixel)}, while at a smaller scale, $z=2$, a map tile is four times larger: 250 meters wide \red{(0.98 meters per pixel)}. Using multiple scales allows large scale tiles to provide scale-space synchronization and small scale tiles to provide high detail \red{; however, they} must be consistent.

In the case of our interactive \red{exploration} system, we can achieve \redr{interactive exploration}{the necessary speed} by limiting the number of tile evaluations \red{to balance speed against} scale or detail. For example, we do not need to generate small scale tiles if the user is viewing a large area. However, small and large \red{scales} must be consistent -- the same areas must have the same content and style. To achieve this scale-space continuity, each \maptosat sub-network is conditioned on a \emph{guidance} subsection of \red{the} larger scale \redr{image}{tile}.

The input to \maptosat at each scale is a map label image and an optional 3-channel color guidance image which is only applied when scale level $z > 1$. The color guidance image is from a larger scale in our hierarchy satellite image that provides color conditioning to ensure consistency between \red{scales}. The output of each $\maptosat^z$ sub-network is a \reds{generated }RGB satellite image consistent with both the map images and the larger-scale color guidance (if provided). The map image, color guidance, and output satellite image have the same resolution\reds{ (256 $\times$ 256 pixels)}.

The principle issue when adapting a CNN generator architecture to an arbitrary resolution is avoiding the memory requirements of large layers.
Using the \seams network we can refactor the network into sub-networks, $\maptosat^z$, \red{for each scale}\reds{for} $z \in 1...z^{max}$ (see Figure~\ref{fig:zoom_arch}). By cropping each sub-network's input, the memory and data requirements are reduced considerably.
With this cropped \reds{sub-module }architecture we are able to train the whole network end-to-end when required, but each sub-network remains independent \red{when evaluated. This} sub-network independence allows portions of scale-space to be evaluated to accelerate our interactive editor.

\reds{old paragraph largely rewritten:}
\red{To implement this refactoring, a spatial crop from 256 $\times$ 256 pixels to 64 $\times$ 64 pixels is inserted between each adjacent sub-network (see Figure~\ref{fig:zoom_arch}); this provides a $4\times$ decrease in scale per sub-network. A 3 $\times$ 3 convolution layer to 512 channels is positioned at the start of each sub-network to preserve the original SPADE network architecture as much as possible -- we found this arrangement did not impact generation quality.}

The inputs to each sub-network are map label images \red{with} optional color guidance images from the previous (larger) scale \red{for sub-networks $z>1$}. The color guidance images coordinate style and content consistency between sub-networks at scales $\maptosat^n$ and $\maptosat^{(n+1)}$. Experimentally, we observe that the SPADE network takes \redr{most of}{the majority of} its guidance from the \reds{map }label image which overpowers the spatial convolutional input guidance. To compensate, we append additional color guidance to the semantic label channel in the spatially-adaptive denormalization \red{blocks} (Figure~\ref{fig:spade_block}). We found that this approach better preserves color guidance information. The color guidance image is blurred with a Gaussian filter ($\sigma = 8$ at 256 $\times$ 256 pixels) to only keep the color information and reduce other high frequency features.  The spatial input to the largest scale $\maptosat^1$ at training time is from an encoder that is trained to create a latent encoding, $e$, from a satellite image; \red{at test time} we use a normally distributed $e$ \red{to create variety over our results}.

The complete \maptosat network is a streamlined version of a high-resolution SPADE with considerably lower memory usage. For instance, the on-disk weights of a full SPADE network trained \redr{by}{at} 256 $\times$ 256 pixels\reds{ image} are 436.9MB \reds{ on disk}\red{ while the weights for a cropped sub-network, $\maptosat^{z>1}$, are 31.5MB}. \red{For a 8,192 $\times$ 8,192 pixels image the full SPADE GPU memory requirement exceeds 32GB, while the factored weights for the two networks in the \name system are less that 12GB}.

Real data is used to train each of the \maptosat sub-networks independently. End-to-end fine-tuning over the whole network results distributions were sufficiently close to realistic data that at test-time these networks would accept synthetic inputs from the previous \maptosat when $z>1$. 
The training data of sub-network $\maptosat^n$ is semantic label map images and color guidance images generated by scale level $\maptosat^{n-1}$. We use around 3,000 images to train the sub-networks. We implement our pipeline with $z^{max}$ of 3 and train every $\maptosat^z$ network for 70 epochs.

\subsection{Complete \name Pipeline}

\begin{figure}[h]
    \centering
    \def\svgwidth{\linewidth}
\begingroup%
  \makeatletter%
  \providecommand\color[2][]{%
    \errmessage{(Inkscape) Color is used for the text in Inkscape, but the package 'color.sty' is not loaded}%
    \renewcommand\color[2][]{}%
  }%
  \providecommand\transparent[1]{%
    \errmessage{(Inkscape) Transparency is used (non-zero) for the text in Inkscape, but the package 'transparent.sty' is not loaded}%
    \renewcommand\transparent[1]{}%
  }%
  \providecommand\rotatebox[2]{#2}%
  \newcommand*\fsize{\dimexpr\f@size pt\relax}%
  \newcommand*\lineheight[1]{\fontsize{\fsize}{#1\fsize}\selectfont}%
  \ifx\svgwidth\undefined%
    \setlength{\unitlength}{696.5931954bp}%
    \ifx\svgscale\undefined%
      \relax%
    \else%
      \setlength{\unitlength}{\unitlength * \real{\svgscale}}%
    \fi%
  \else%
    \setlength{\unitlength}{\svgwidth}%
  \fi%
  \global\let\svgwidth\undefined%
  \global\let\svgscale\undefined%
  \makeatother%
  \begin{picture}(1,0.50170696)%
    \lineheight{1}%
    \setlength\tabcolsep{0pt}%
    \put(0,0){\includegraphics[width=\unitlength,page=1]{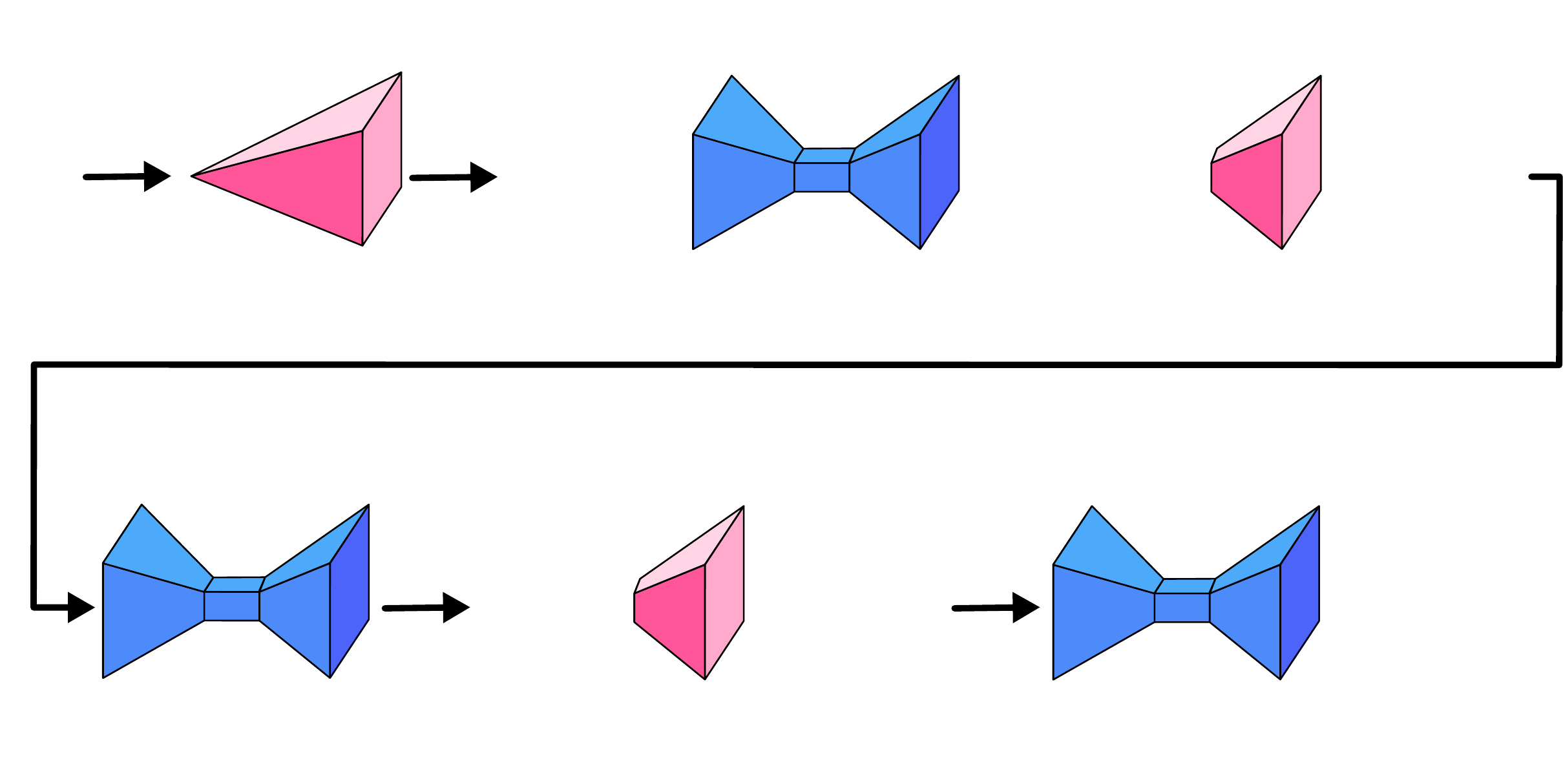}}%
    \put(0.18501175,0.31318868){\color[rgb]{0,0,0}\makebox(0,0)[t]{\lineheight{1.25}\smash{\begin{tabular}[t]{c}$\maptosat^1$\end{tabular}}}}%
    \put(0.81211033,0.30974514){\color[rgb]{0,0,0}\makebox(0,0)[t]{\lineheight{1.25}\smash{\begin{tabular}[t]{c}$\maptosat^2$\end{tabular}}}}%
    \put(0.44051517,0.03368791){\color[rgb]{0,0,0}\makebox(0,0)[t]{\lineheight{1.25}\smash{\begin{tabular}[t]{c}$\maptosat^3$\end{tabular}}}}%
    \put(0.52687189,0.31093908){\color[rgb]{0,0,0}\makebox(0,0)[t]{\lineheight{1.25}\smash{\begin{tabular}[t]{c}$\seams^1$\end{tabular}}}}%
    \put(0.14761938,0.03482362){\color[rgb]{0,0,0}\makebox(0,0)[t]{\lineheight{1.25}\smash{\begin{tabular}[t]{c}$\seams^2$\end{tabular}}}}%
    \put(0.75567992,0.03461048){\color[rgb]{0,0,0}\makebox(0,0)[t]{\lineheight{1.25}\smash{\begin{tabular}[t]{c}$\seams^3$\end{tabular}}}}%
    \put(0.03466899,0.3796482){\color[rgb]{0,0,0}\makebox(0,0)[t]{\lineheight{1.25}\smash{\begin{tabular}[t]{c}$e$\end{tabular}}}}%
    \put(0,0){\includegraphics[width=\unitlength,page=2]{scit.pdf}}%
  \end{picture}%
\endgroup%

    \caption{\name combines a series of \maptosat and \seams networks. Here we show a 3-scale ($z^{max}=3$) pipeline. \maptosat sub-networks generate  satellite image tiles which are processed by the \seams networks to remove spatial discontinuities.}
    \label{fig:scit}
\end{figure}

\begin{figure}[h]
    \centering
    \def\svgwidth{\linewidth}
    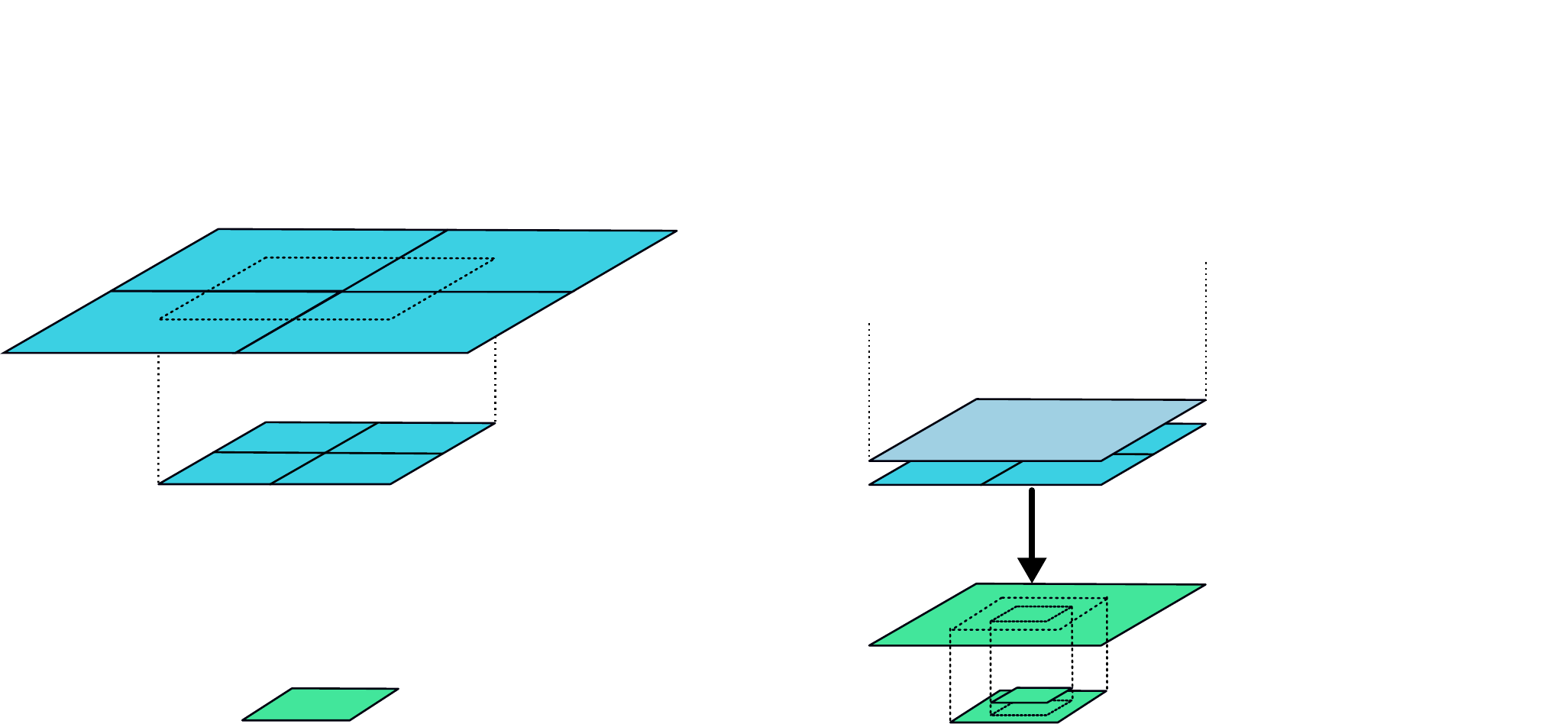
    \caption{The geometry of \name tiles. Left, right columns: network outputs at scales $z=n$ and $z=n+1$. Scale-space continuity is provided by passing a blurred crop of $y^n$ (green bottom left) to $\maptosat^{n+1}$ (cyan mid-right). }
    \label{fig:geometry}
\end{figure}

Our entire system is a combination of the \seams and \maptosat networks (Figure~\ref{fig:overview}) over the scale levels. The architecture is shown in Figure~\ref{fig:scit} and the tile geometry in Figure \ref{fig:geometry}. At the $z^1$ scale level, we use $\maptosat^1$ with random latent encoding $e$, but without color guidance. The satellite images tiles $s^1_i$ and $t^1_i$ are generated and processed by the $\seams^1$ network for spatial continuity, resulting in $y'^1_i$. The set of seamless tiles, $y'^1_*$, are stitched together to create the final image at scale $z^1$; we crop the this result to use as the color guidance of $z^2$. This process continues over all scales $z \in 1...z^{max}$ to \red{generate satellite images of different scales over an area.}

Because our method \red{generates tiles and stitches} them together to \redr{get}{create} huge images, we can generate satellite images of arbitrary size \redr{for}{from} cartographic map data. Furthermore, we can enforce scale continuity over \redr{arbitrary}{a wide range of} scales.
\red{Another advantage of our \name pipeline is that the two independent network architectures and independent networks at each scale can be used for custom applications; for example, the $z^2$ \seams network may be used to remove seams in real-world satellite images, or we may insert additional sub-modules to create satellite images with smaller scales.}

\red{
The general complexity of the entire SSS pipeline depends on the scale level, $z=k$, and the linear scale factor, $f$, used between each sub-network. For example, a scale factor of $f = 4$ between each sub-network implies that the smaller scale sub-network has $f^2 = 16$ tiles covering the area of a single larger scale tile. To generate one scale, all larger scales must also be generated. 
The worst-case number of \maptosat evaluations, assuming we generate $4$ tiles at $z=k$ is:
    \begin{equation}
        \sum_{i=1}^{k} 4/(f^2)^{i-1}
    \end{equation}
and worst-case number of \seams evaluations is:
    \begin{equation}
        \sum_{i=1}^{k} (\sqrt{4/(f^2)^{i-1}}-1)^2        
    \end{equation}
}

\reds{Therefore, generating n tiles at scale $z$ in our system needs $ n/16^0 + n/16^1 + ... + n/16^{z-1} $ \maptosat evaluations and $ (\sqrt{n/16^0}-1)^2 + (\sqrt{n/16^1}-1)^2 + ... + (\sqrt{n/16^{z-1}}-1)^2 $ \seams evaluations. }

\section{Results}
\label{sec:results}

\red{In the following results we use \name with $z^{max}=3$ and a scale factor of $f=4$ to create} a 512$\times$512 pixels image at $z^1$, a 2,048$\times$2,048 pixels image at $z^2$, and a 8,192$\times$8,192 pixels image at $z^3$. 

\begin{figure*}[h]
    \centering
    \def\svgwidth{\linewidth}
    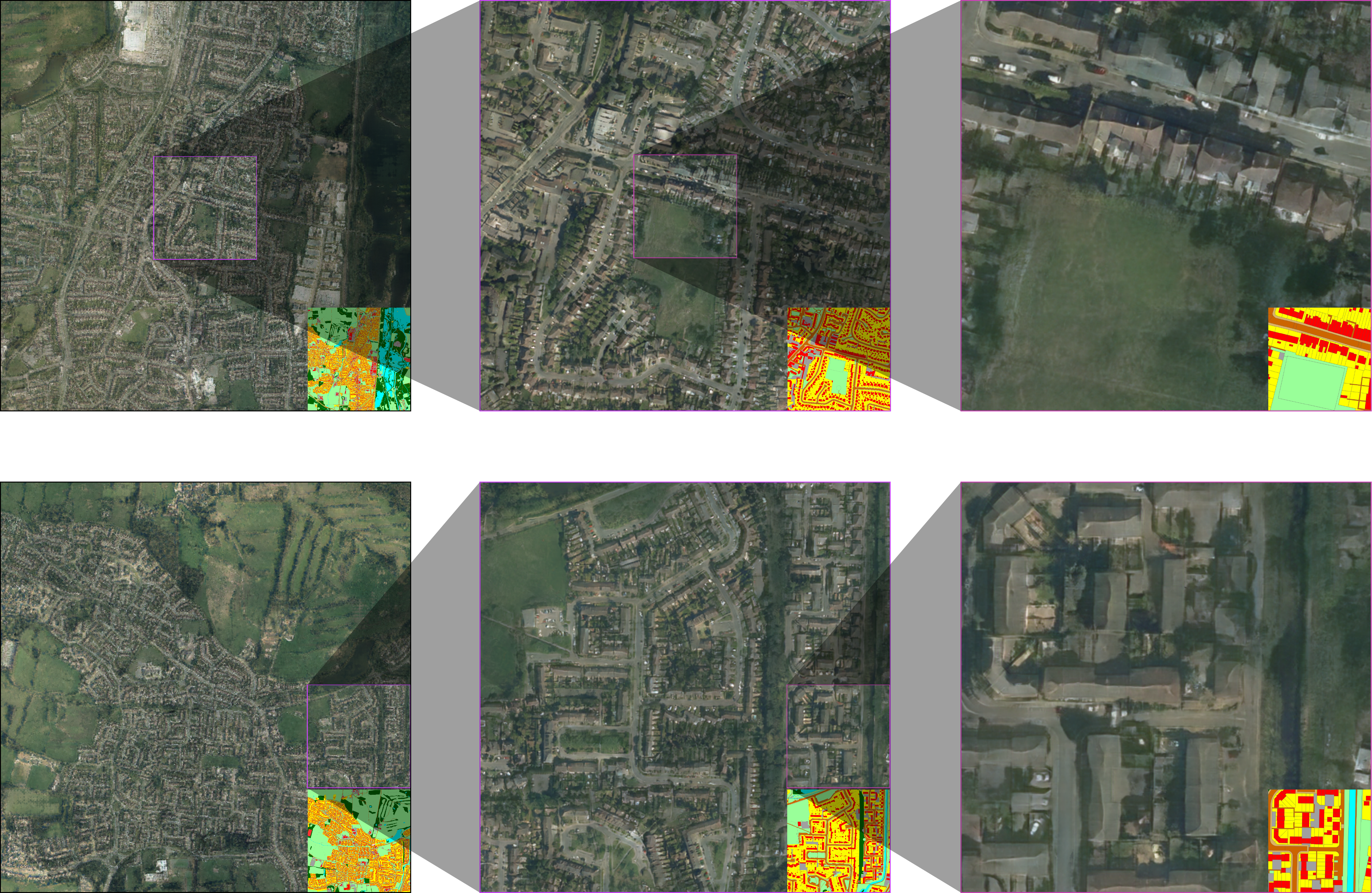
    \caption{Results from the \name pipeline. Rows: examples of generated satellite images from real-world map data. Columns: left to right show scales $z=1,2,3$. Inset are the corresponding map inputs. }
    \label{fig:results}
\end{figure*}

\begin{figure*}[h]
    \centering
    \def\svgwidth{\linewidth}
    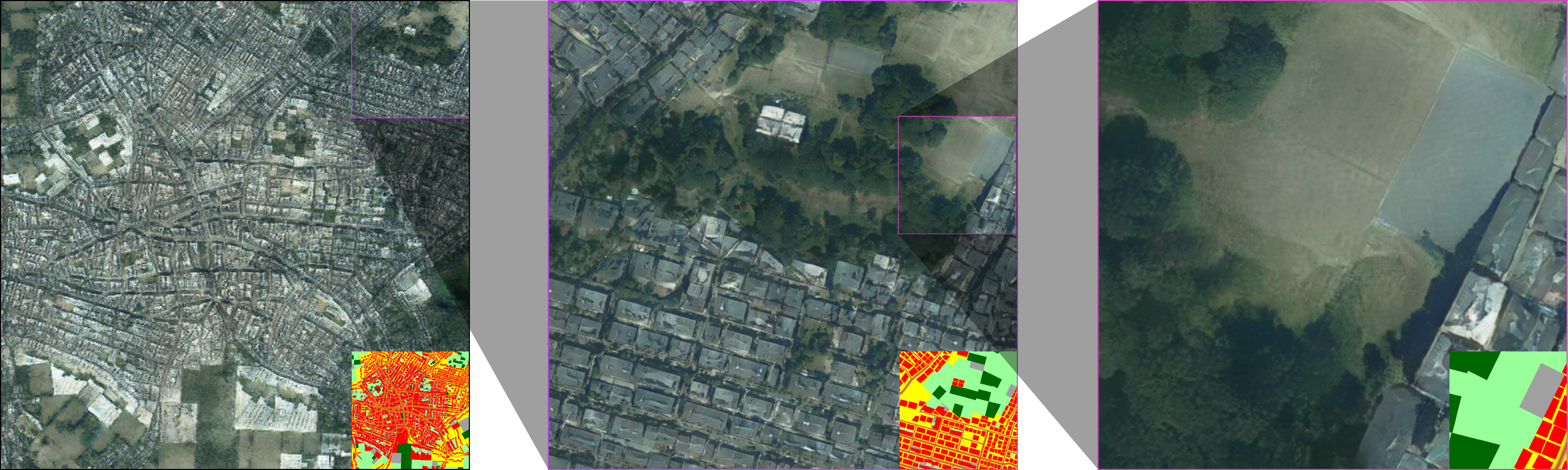
    \caption{Result of our system on synthetic procedurally generated map data.}
    \label{fig:p_result}
\end{figure*}

\begin{figure}[ht]
    \centering
    \def\svgwidth{\linewidth}
    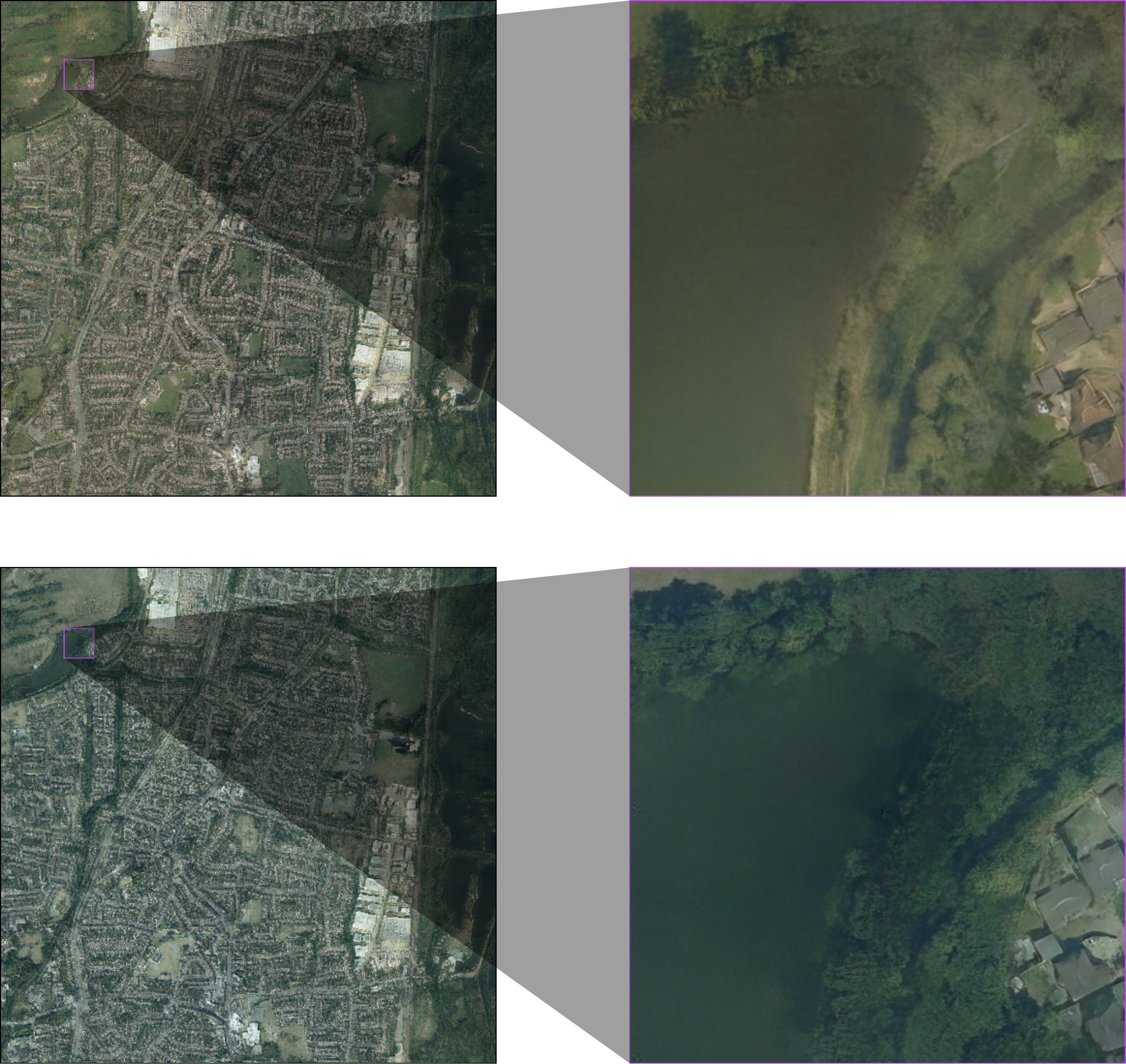
    \caption{Style generation of $z^1$ (left) and $z^3$ (right) images in a 3 scale level \name system. Each row has a different latent encoding, $e$, input to $\maptosat^1$, giving different styles which remain consistent throughout the scale levels.}
    \label{fig:consistency}
\end{figure}

\red{Figure~\ref{fig:results}} shows the output of our technique for\reds{input areas on} real-world map \maps; further results are given in our additional materials (see folder \emph{sss\_8k\_images\slash}). Figure~\ref{fig:p_result} shows the output of our technique for\reds{ input areas on} synthetic map labels generated using a procedural modeling system. \reds{Additional full-resolution results are provided in additional materials.}Figure~\ref{fig:consistency} illustrates the style continuity of \name pipeline.

\begin{figure}
	\centering
	\includegraphics[width=1.0\columnwidth]{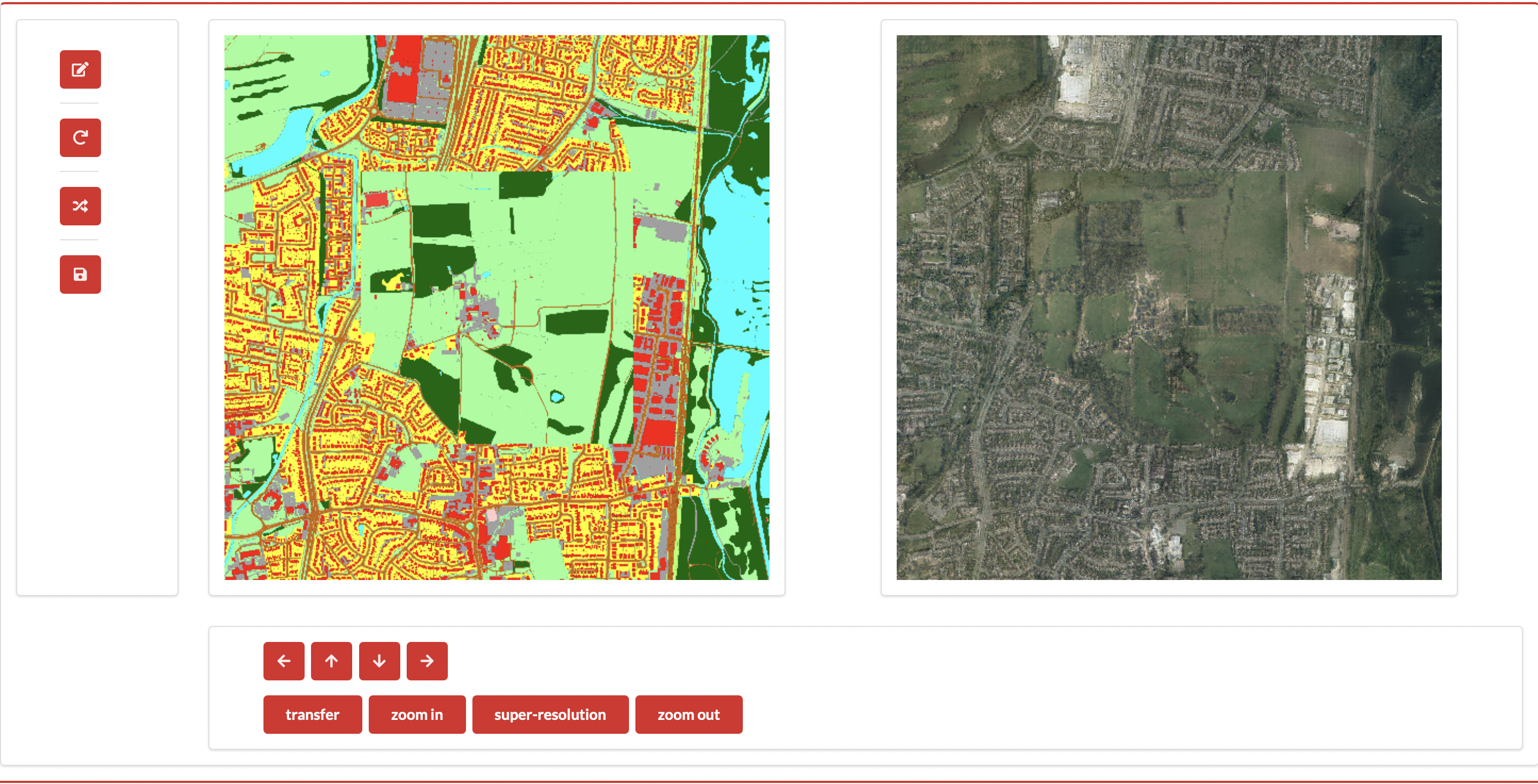}
	\caption{User interface for our interactive system. In this figure, we show a result of generated satellite image (right) by corresponding map (left). Here the user has edited the map by rotating the central tile. Our system is able to generate realistic results even with the discontinuities in the map data. }
	\label{fig:ui}
\end{figure}

Finally, our interactive system is introduced in figure~\ref{fig:ui}. We allow users to move around in the map, \red{zoom in or out}, \redr{ coordinate system provided in advance. Users can}{and} generate corresponding satellite images of different styles.\reds{and view these images in different scale in scale-space} We also provide some operations to modify the map, such as randomly replacing a \redr{discontinuous}{single} tile in the map \redr{image}{input}. All networks (\maptosat and \seams) for 3 scales can be loaded into the 12GB memory of a Nvidia Titan V graphics card. Generation times for a 512 $\times$ 512 pixels window vary between 2.4 and 3.42 seconds depending on the scale. Smaller scales require additional network evaluations.

\section{Evaluation}

\subsection{Methodology}
Evaluating the quality of composite synthesized images for neural networks is a difficult problem; there is no canonical method to assess the performance of image stitching. Traditional edge detection technology, such as Sobel filter detection and Canny edge detection, can be used for seams detection but are not optimized for tile-based systems where the seam locations can be anticipated. In addition, these algorithms output images with edges rather than comparable values.

To evaluate our \seams network results, we use two methods. First, we design a qualitative user study survey. We ask users to identify the best system for removal of seams between tiles to examine the effectiveness of our \seams network against other algorithms\reds{ that can remove seams}. Second, we introduce a novel algorithm, MoT (Mean over Tiles), to \redr{detect}{measure the ability to} remove seams in \redr{our generated results}{images}. We evaluate against the following baselines\reds{that are from prior works}: traditional image blending with a soft mask, Graphcut\cite{kwatra2003graphcut}, Generative Image Inpainting\cite{yu2019free}, and TileGAN\cite{fruhstuck2019tilegan}. 

\begin{figure}
    \centering
    \def\svgwidth{\linewidth}
\begingroup%
  \makeatletter%
  \providecommand\color[2][]{%
    \errmessage{(Inkscape) Color is used for the text in Inkscape, but the package 'color.sty' is not loaded}%
    \renewcommand\color[2][]{}%
  }%
  \providecommand\transparent[1]{%
    \errmessage{(Inkscape) Transparency is used (non-zero) for the text in Inkscape, but the package 'transparent.sty' is not loaded}%
    \renewcommand\transparent[1]{}%
  }%
  \providecommand\rotatebox[2]{#2}%
  \newcommand*\fsize{\dimexpr\f@size pt\relax}%
  \newcommand*\lineheight[1]{\fontsize{\fsize}{#1\fsize}\selectfont}%
  \ifx\svgwidth\undefined%
    \setlength{\unitlength}{637.19556914bp}%
    \ifx\svgscale\undefined%
      \relax%
    \else%
      \setlength{\unitlength}{\unitlength * \real{\svgscale}}%
    \fi%
  \else%
    \setlength{\unitlength}{\svgwidth}%
  \fi%
  \global\let\svgwidth\undefined%
  \global\let\svgscale\undefined%
  \makeatother%
  \begin{picture}(1,0.73011835)%
    \lineheight{1}%
    \setlength\tabcolsep{0pt}%
    \put(0,0){\includegraphics[width=\unitlength,page=1]{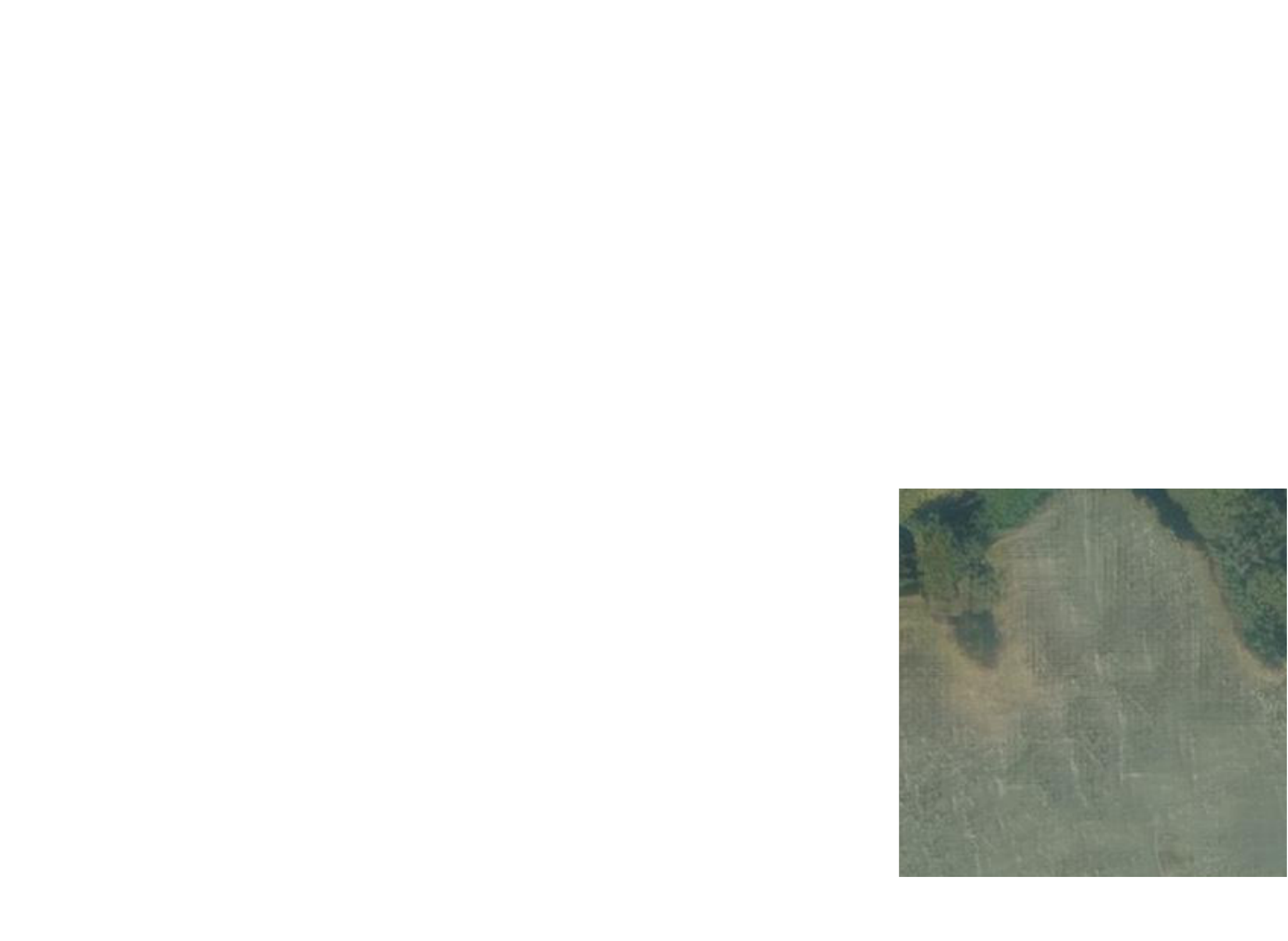}}%
    \put(0.48674489,0.38876818){\color[rgb]{0,0,0}\makebox(0,0)[t]{\lineheight{1.25}\smash{\begin{tabular}[t]{c}(b)\end{tabular}}}}%
    \put(0.13908916,0.38732826){\color[rgb]{0,0,0}\makebox(0,0)[t]{\lineheight{1.25}\smash{\begin{tabular}[t]{c}(a)\end{tabular}}}}%
    \put(0.1396092,0.00947137){\color[rgb]{0,0,0}\makebox(0,0)[t]{\lineheight{1.25}\smash{\begin{tabular}[t]{c}(d)\end{tabular}}}}%
    \put(0,0){\includegraphics[width=\unitlength,page=2]{bl.pdf}}%
    \put(0.83140698,0.38744773){\color[rgb]{0,0,0}\makebox(0,0)[t]{\lineheight{1.25}\smash{\begin{tabular}[t]{c}(c)\end{tabular}}}}%
    \put(0.49109716,0.00976462){\color[rgb]{0,0,0}\makebox(0,0)[t]{\lineheight{1.25}\smash{\begin{tabular}[t]{c}(e)\end{tabular}}}}%
    \put(0.82961595,0.01103045){\color[rgb]{0,0,0}\makebox(0,0)[t]{\lineheight{1.25}\smash{\begin{tabular}[t]{c}(f)\end{tabular}}}}%
    \put(0,0){\includegraphics[width=\unitlength,page=3]{bl.pdf}}%
  \end{picture}%
\endgroup%

    \caption{This figure illustrates results from selected baseline techniques. (a) $t$ from our \maptosat network; (b) the output of our \seams network; (b, inset) the calculated mask, $m$; (c) the result of fixed-mask  (c, inset) image blending. (d) the Graphcut algorithm calculates four minimum energy seams within margins (d, inset); (e) Generative Image Inpainting is not conditioned on map data, but is used to inpaint a diamond area (e, inset); (f) TileGAN conditioned on a lower-resolution satellite image.}
    \label{fig:bl}
\end{figure}

\reds{Before we start the evaluation, we implement some algorithms and models that can finish the similar task. We use these as baselines for comparison. }

For traditional image blending, we use $s$ as input and blend $s$ with $t$ by a soft mask shown in the right bottom corner of Figure~\ref{fig:bl} (c) to obtain an image without seams. Graphcut removes seams by splicing the center part of $t$ into $s$ selecting a cut through a buffer area. We use a small area, 10 $\times$ 10 pixels, shown in the right bottom corner of Figure~\ref{fig:bl} (d) as \red{our} buffer area. \redr{Then w}We train the Generative Image Inpainting model, which \redr{it is a neural network using}{uses} gated convolutions\cite{yu2019free} \redr{to finish}{for the} image inpainting task; it is trained with our generated seams dataset $s$ and the diamond shape mask shown in the right bottom corner of Figure~\ref{fig:bl} (e). This is to let the model learn to generate the center diamond area which covers \red{the} seams\reds{ area}. Generative Image Inpainting does synthesize reasonable results, but it often changes the original content structure in $s$ as it is not conditioned on map data. Last, we train PGGAN~\cite{karras2017progressive} with ground-truth satellite images of 256$\times$256 \red{pixels} resolution for TileGAN. TileGAN \reds{is }can generate large seamless images \reds{and it does not need to be trained. however, it needs a pre-trained PGGAN model and}by searching the latent space of the pre-trained PGGAN model to create tiles that are \redr{used for stitching}{stitched} into a large image. It is not strictly an image-to-image translation system, but can take a small, low-resolution, image as guidance to generate large, high-resolution, images. 

Figure~\ref{fig:bl} shows selected representative output tiles for each of \redr{these approaches}{the baselines}. Traditional image blending in Figure~\ref{fig:bl} (c) \redr{finishes}{does} a good job generally, but since it use\red{s} a fixed mask, the rough shape of the mask can \red{sometimes} be observed\reds{ carefully}. This is more obvious when we apply image blending on green fields. Figure~\ref{fig:bl} (d) is created by \red{the} Graphcut algorithm, \red{which clearly shows visible seams}.\reds{ There are clearly splice traces in this one.} Figure~\ref{fig:bl} (e) is generated by Generative Image Inpainting; \redr{showing quite}{it does synthesize} good results, \redr{but}{however,} it changes the original structure of Figure~\ref{fig:bl} (a); traces of the diamond \reds{shape }mask can also be observed\reds{ if looking carefully}. 

Figure~\ref{fig:bl} (f) is a tile from a large image \redr{systhesized}{synthesized} by TileGAN; \reds{We generate }an 8,320 $\times$ 8,320 pixel image \red{is} guided by a\reds{n area's} low-resolution image \red{which is cropped into} the correspond\red{ing} tile\reds{ image}. TileGAN creates an overly ``blobby'' result \red{ with poor large scale features}\reds{ with poor street reconstruction}. 
While TileGAN is true to the guidance image, \redr{it}{the guidance} is too low resolution to effectively guide the generation of crisp man-made features \red{such as streets}.
When TileGAN is trained on satellite images of urban areas, the results can be poor because of the difficulty in searching for a latent encoding which matches large, \redr{precise}{sharp} features over tile boundaries. This is illustrated in the additional materials file \emph{baselines\_8k\_images\slash tilegan.jpg}. We\reds{ also} provide large area results (8k images) of all \reds{these }baselines \reds{above }in \emph{baselines\_8k\_images\slash}. 

We also evaluate against the following variants and later use these \reds{variants }to form an ablation study: photographic ground-truth, ground-truth with seams (created by flipping each tile along the diagonal), original SPADE\cite{pix2pix} with no \seams networks, our \maptosat results without \seams network, and our results which are generated by procedural map data.

\begin{figure}[h]
    \def\svgwidth{\linewidth}
    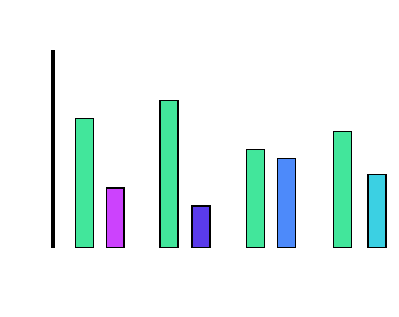
    \caption{User study results. We show results of \name (our approach), Graphcut,  Generative Image Inpainting (GII) and $t$ (\maptosat output, without \seams). In all scenarios, users rate our seam-removal as more natural than alternatives.}
    \label{fig:user}
\end{figure}

\textbf{The User study survey} evaluates user's perceived quality of seam removal. We designed a user questionnaire which shows an image with seams, and ask which of two techniques remove the seams in the most natural way. We compare against 4 baselines: Graphcut, Generative Image Inpainting, image blending with a soft mask, and the dataset $t$ without seams that we use for training. Comparison trials for one baseline are performed over 10 pairs of images. We do not include TileGAN in the survey because it does not support generating satellite imagery of \redr{the}{a} specifi\redr{ed}{c} area. We asked a total 20 participants ``Which of these two images below removes the seams in the most natural way?'' (see \redr{Appendix for example screens}{\emph{misc\slash user\_study.png} in additional materials}). We also collected information on participant's digital image editing experience. Among our participants, 55$\%$ process digital images every month, while the remaining 45$\%$ have no relevant experience.

Figure \ref{fig:user} shows the results of our user study. \name removes seams in a more natural \redr{seam-removal}{way} than all the baselines in the majority of trials. Our advantage is particularly strong against Graphcut and Generative Image Inpainting. We believe this is because Graphcut cannot complete the stitching task well over a limited buffer area, and Generative Image Inpainting removes seams while changing the original content of the image.
We found it surprising that when our results are compared with $t$, we still have an advantage. \redr{This may be because many users think that although $t$ does not have seams at all, its outside area may have changed. }{This may be because the tile generation near the border ($t$) may be less accurate, so showing more images from the center of adjacent tiles ($y'$) may create more realistic results.}

\begin{figure*}
    \def\svgwidth{\linewidth}
    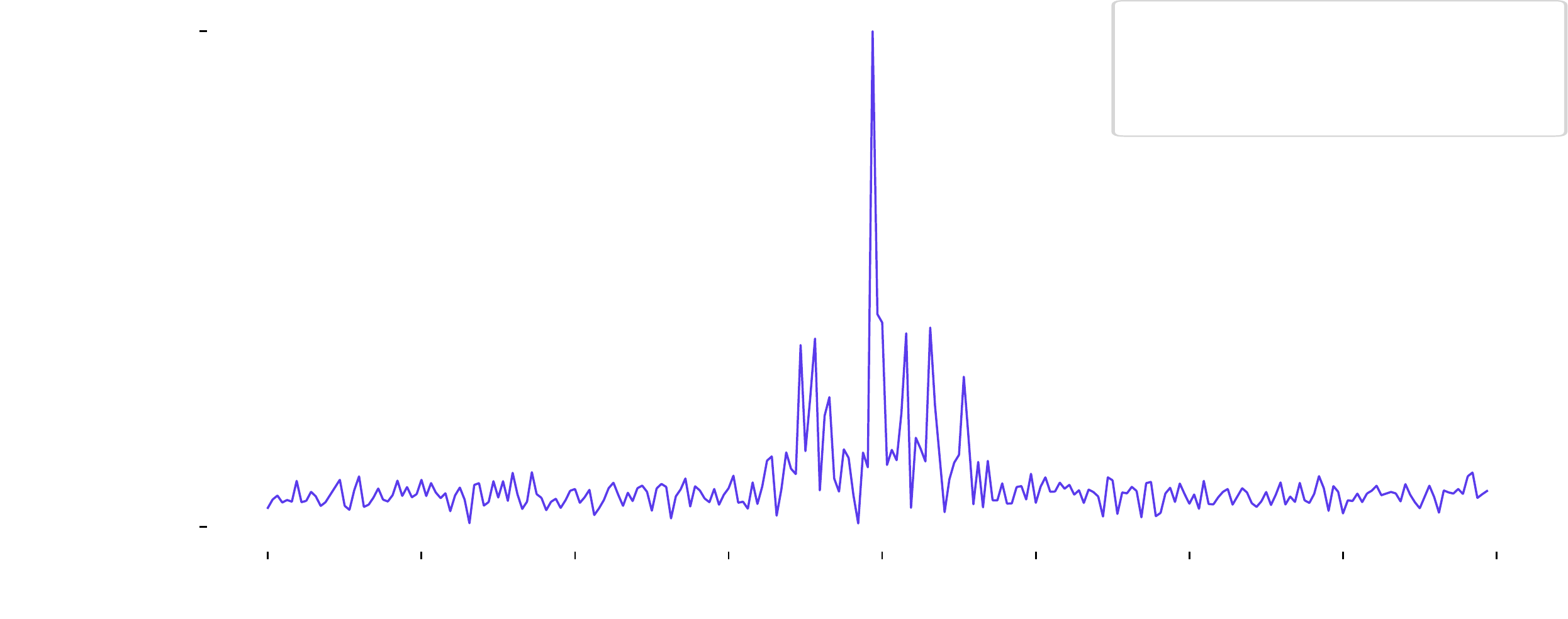
    \caption{Selected MoT algorithm results: difference value response for ground-truth, ground-truth with seams, our \maptosat with no \seams and our whole pipeline. Images with strong vertical and horizontal seams have a strong response on the tile boundary at 128 pixels. \reds{The image containing seams reflects in this figure that their line does have a peak (we consider a place that is significantly higher than twice the second highest value as a peak).}}
    \label{fig:eval}
\end{figure*}

\textbf{The MoT algorithm} compares differences between horizontally and vertically adjacent pixels over generated images to highlight any sudden color differences at \red{known} tile boundaries. 

\red{To compute the MoT response we} firstly \reds{we }crop the input image by removing ($tileWidth/2$) pixels from each edge. Let $C_a$ be all pixels in columns $\{0+a, tileWidth+a, tileWidth*2+a$,...,$imageWidth-tileWidth+a\}$ from a cropped image. Similarly $R_a$ for all rows. The MoT response at pixel position, $x,y$, is the absolute difference of the averages of differences of adjacent pixels:
    \begin{equation}
       D(x,y) = |\overline{C_x} - \overline{C_{x+1}}| + |\overline{R_y} - \overline{R_{y+1}}|
    \end{equation}
    
We use $tileWidth = 256$ to match our generated tile size, $x,y \in 0...tileWidth-1$, and a 8,192 $\times$ 8,192 pixel image to calculate the MoT values.
    
Figure \ref{fig:eval} shows the results of the MoT algorithm. The horizontal axis shows discontinuities at each pixel. \red{The \maptosat results show}\reds{We can see that before we run \seams network on our results there is} an obvious response peak at tile boundaries, \red{while}\reds{and } after \red{the} \seams network our results do not have this peak. This illustrates that our network has learned to remove the seams over the image. This algorithm also works for other baselines and variants; table~\ref{table:mot} shows the absolute difference between of $max(D(x,y))$ and $mean(D(x,y))$ over $x,y$ for the baselines and variants. When the difference value is greater than 0.002 a sharp vertical or horizontal seam is present.

\begin{table}
\centering
\begin{tabular}{|l|l|}
\hline
Model, Baseline, variant&difference value\\
\hline
Image blending with a soft mask&0.000341\\
\hline
Graphcut&0.000344\\
\hline
Generative Image Inpainting&0.001911\\
\hline
TileGAN&0.001336\\
\hline
Ground-truth&0.000291\\
\hline
Ground-truth with seams&0.003214\\
\hline
Original SPADE without \seams&0.002723\\
\hline
The \maptosat without \seams &0.004724\\
\hline
The \name whole pipeline&0.000730\\
\hline
\makecell[l]{The \name whole pipeline \\with procedural map data}&0.000845\\
\hline
\end{tabular}
\caption{The difference between max and mean values from our MoT algorithm. We found that images with difference values greater than 0.002 contained obvious seams. }
\label{table:mot}
\end{table}

\subsection{Ablation Study}
\label{sec:ablation}

We use variants' results from \redr{ last}{the previous} section and some new variants to present a visual ablation study. First, we create \redr{one}{an} image of 512$\times$512 \red{pixels} resolution at scale $z^1$ and \red{upscale it with simple Lanczos interpolation to 8,192 $\times$ 8,192 pixels}. \redr{We can see from the results that his image}{Such an approach} loses \reds{all }its detail, \red{appearing very soft}\reds{ and end up with a very fuzzy look} (see additional materials, \emph{variants\_8k\_images\slash large\_scale\_only.jpg}). Second, we create an 8k image only at $z^3$, \red{without color guidance,} and \red{note that} this image looks very homogeneous. For example, most natural environmental areas are covered with \red{the} same color and often have similar patterns (see \emph{variants\_8k\_images\slash small\_scale\_only.jpg}). Third, we generate an 8k image with \maptosat, but no \seams network (see  \emph{variants\_8k\_images\slash no\_sat2tile\_network.jpg}). This gives us an \reds{stitched large} image with seams, but because this image is generated using \reds{ the result of the $z^2$ as the} color guidance, the style of each tile also \red{shows scale-space}\reds{reflects the} consistency; it is \redr{still}{also} better than \redr{the}{a} large image stitched together by \red{the} results of SPADE alone (see \emph{baselines\_8k\_images\slash seams.jpg}).

\vspace{-0.5cm}
\section{Limitations}

There are several limitations of our \name pipeline. First, our \maptosat is based on SPADE\cite{park2019semantic} \red{and Pix2Pix\cite{pix2pix} } which means that the quality of the generated satellite images is limited \red{by these architectures}. We also tried Progressive Growing of GANs when we implementing TileGAN for comparison, but there is still a gap between the generated satellite image and the ground-truth satellite image. 
\red {Another limitation is that we only remove the linear seams between adjacent tiles. The \seams network also leaves theoretical singular point discontinuities at the midpoints of each tile edge after blending between $s$ and $t$. We found this had minimal impact on the quality of seam removal in practice, but can envisage an additional network which removes such singular discontinuities.}
\reds{Next, our \seams network performs well especially when working with \maptosat, but there is a theoretic possibility that some points (e.g. in the center of each edge) introduce local artifacts. }Finally, we produce strong results because \name is conditioned on cartographic data, but compare to some baselines which can not use this data fully (e.g., TileGAN), or were not designed for this task (e.g., Graphcut).
\section{Conclusion}
\label{sec:conclusion}

We have proposed the \name pipeline to synthesize large satellite images that are continuous both \red{ spatially and} throughout scale-space\reds{ and spatial-space} at arbitrary size for given cartographic map data while \reds{do }not requir\red{ing large} \reds{huge }memory \redr{usage}{use}. It also supports procedurally generated cartographic maps for entirely novel areas which allow users to create textures on demand. This is achieved by the \maptosat network architecture that can keep style continuity in specified scale-space and \seams network architecture to splice small images seamlessly into one large image. We also implement a interactive system to let users explore of an boundless satellite images generated conditioned on cartographic data.

In future work, we would like to we would like to study the impact of different edge constraint images ($c$) in our \seams network \red{by replacing $c$ with blob-shaped masks} and us\red{ing} other conditioning for \maptosat, such as multi-spectral images, pencil sketches, and low resolution images. Another avenue of exploration is the scale factor of 4 between $z$; we would like to try different values with a perceptual evaluation. In addition, we suspect that using different offsets at different scales would reduce regularity and improve the quality of results. \red{We also want to try to modify the neural network structure used; for example, adding adaptive discriminator augmentation\cite{karras2020training}. }Finally, we want to extend our work to 3D voxel based networks using transformers to synthesize a 3D virtual environment from cartographic map data.

\section{Acknowledgments}

We would like to thank Nvidia Corporation for hardware and Ordnance Survey Mapping\cite{digimap, mastermap} for map data which made this project possible. This work was undertaken on ARC4, part of the High Performance Computing facilities at the University of Leeds, UK. This work made use of the facilities of the N8 Centre of Excellence in Computationally Intensive Research (N8 CIR) provided and funded by the N8 research partnership and EPSRC (Grant No. EP/T022167/1).

\bibliographystyle{eg-alpha-doi}  
\bibliography{bib.bib}

\end{document}